\def\year{2020}\relax
\documentclass[letterpaper]{article} 
\usepackage{aaai20}  
\usepackage{times}  
\usepackage{helvet} 
\usepackage{courier}  
\usepackage[hyphens]{url}  
\usepackage{graphicx} 
\usepackage{graphicx}  

\usepackage{amsmath}
\usepackage{amssymb}
\usepackage{subfigure}
\usepackage{booktabs}
\usepackage{array}
\usepackage{multirow}
\usepackage[export]{adjustbox}
\usepackage{color}
\newcolumntype{L}[1]{>{\raggedright\let\newline\\\arraybackslash\hspace{0pt}}m{#1}}
\newcolumntype{C}[1]{>{\centering\let\newline\\\arraybackslash\hspace{0pt}}m{#1}}
\newcolumntype{R}[1]{>{\raggedleft\let\newline\\\arraybackslash\hspace{0pt}}m{#1}}

\newcommand{\citet}[1]{\citeauthor{#1} \shortcite{#1}}
\newcommand{\citep}{\cite}

\title{DSTC8-AVSD: Multimodal Semantic Transformer Network \\ with Retrieval Style Word Generator}
\Large\author{Hwanhee Lee\textsuperscript{1}, Seunghyun Yoon\textsuperscript{1}, Franck Dernoncourt\textsuperscript{2}, Doo Soon Kim\textsuperscript{2}, Trung Bui\textsuperscript{2} and Kyomin Jung\textsuperscript{1}\\  
\textsuperscript{1}Department of Electrical and Computer Engineering, Seoul National University, Seoul, Korea \\
\textsuperscript{2}Adobe Research, San Jose, CA, USA\\
\{wanted1007, mysmilesh, kjung\}@snu.ac.kr, 
\{franck.dernoncourt, dkim, bui\}@adobe.com}
\begin{document}

\maketitle

\begin{abstract}
Audio Visual Scene-aware Dialog (AVSD) is the task of generating a response for a question with a given scene, video, audio, and the history of previous turns in the dialog. Existing systems for this task employ the transformers or recurrent neural network-based architecture with the encoder-decoder framework. Even though these techniques show superior performance for this task, they have significant limitations: the model easily overfits only to memorize the grammatical patterns; the model follows the prior distribution of the vocabularies in a dataset.
To alleviate the problems, we propose a Multimodal Semantic Transformer Network. It employs a transformer-based architecture with an attention-based word embedding layer that generates words by querying word embeddings. With this design, our model keeps considering the meaning of the words at the generation stage. The empirical results demonstrate the superiority of our proposed model that outperforms most of the previous works for the AVSD task.

\end{abstract}

\section{Introduction}

Recently, multimodal tasks such as Visual Question Answering (VQA) and visual dialog have attracted much attention in the field of artificial intelligence~\cite{goyal2017making,das2017visual}. 
This task is considered difficult to tackle since incorporating various types of input (i.e., audio, video, and text) requires different techniques to be jointly applied.
One such example, Audio Visual Scene-aware Dialog (AVSD), is the task of answering a question for a video clip and its corresponding dialogue history~\cite{alamri2019audio,kim2019eighth}. The AVSD is considered a more challenging task than a visual dialog task because the model needs to deal with additional audio context, which is consists of sequential information. 
Furthermore, it needs to recognize the history of dialog along with the visual and acoustic data for accurately answering the question.
Previous works~\cite{hori2019joint,le2019multimodal,schwartz2019simple} tackled the AVSD task with the sequence-to-sequence framework~\cite{sutskever2014sequence}, which is composed of encoder and decoder architecture. Although sequence-to-sequence approaches show successful performance in various sentence generation tasks, the technique shows major limitation when applying to multimodal question answering tasks: it easily overfits only to memorize the grammatical patterns and the prior distribution of the vocabularies in a dataset; it fails to model the semantic information in the multimodal dataset.

For the AVSD task, this problem can be exacerbated since the pattern of the question in the dataset is constructed in a straightforward form. For example, for the question ``What color is the door," the model quickly learns the answer pattern, ``the door is [any color] in color," regardless of the actual color information of the door. In other words, most tokens for generating the answer will be selected without considering what the door's real color is. With this approach, the output of the model receives a high automatic evaluation score (i.e., BLEU score) for a sentence without incorporating information from other modalities (i.e., video and audio). 

To alleviate the problem, we proposed multimodal semantic
transformer networks (\textbf{MSTN}) that employ an attention-based word embedding layer, which is inspired by~\citet{ma2018query}'s work, in the decoder part of the Transformer architecture~\cite{vaswani2017attention}.
With the proposed architecture, the \textbf{MSTN} keeps considering the meaning of the words at the generation stage.
Experimental results show that our proposed method surpasses all of the baseline models.

\section{Related Work}

{\bf Multimodal Transformer Network}
Recently~\cite{le2019multimodal} proposed Multimodal Transformer Networks (MTN), which is a state-of-the-art system for AVSD tasks. The MTN is a transformer-based encoder-decoder framework that has several attention blocks to incorporate multiple modalities such as video, audio, and text. The MTN is composed of three major components: encoder layers, decoder layers, and query-aware auto-encoders. Encoder layers encode each input by layer normalization and positional encoding. Decoder layers generate the target sequences and use the multi-head attention mechanism to incorporate information from multiple modalities. Finally, Auto-encoder layers encode video features with query-aware attention and then tries to reconstruct the queries to learn the attention in an unsupervised way. Our model is also based on the transformer encoder-decoder framework, but our model does not use auto-encoder loss and have more simple architecture.

\vspace*{2mm}
\noindent
{\bf Word Embedding Attention Network}
Sequence-to-sequence architecture generally memorizes word patterns in the training dataset other than understanding the meaning of the word itself. Hence, the generated results from the model may not reflect the semantic information of words. \cite{ma2018query} point the main reason for this phenomenon is the output layer of the decoder that consists of a linear transformation and a softmax operation. To solve this problem, \citet{ma2018query} proposes Word Embedding Attention Network (WEAN), which is a retrieval style word generator compared to the linear projection layer in general sequence-to-sequence architecture. By substituting the linear projection layer with WEAN, the model produces words by querying word embeddings, considering word meanings other than memorizing the sequence pattern. We adopt this retrieval style word generator to this task.

\section{Model}
\begin{figure}[t]
\centering
\includegraphics[scale=0.3]{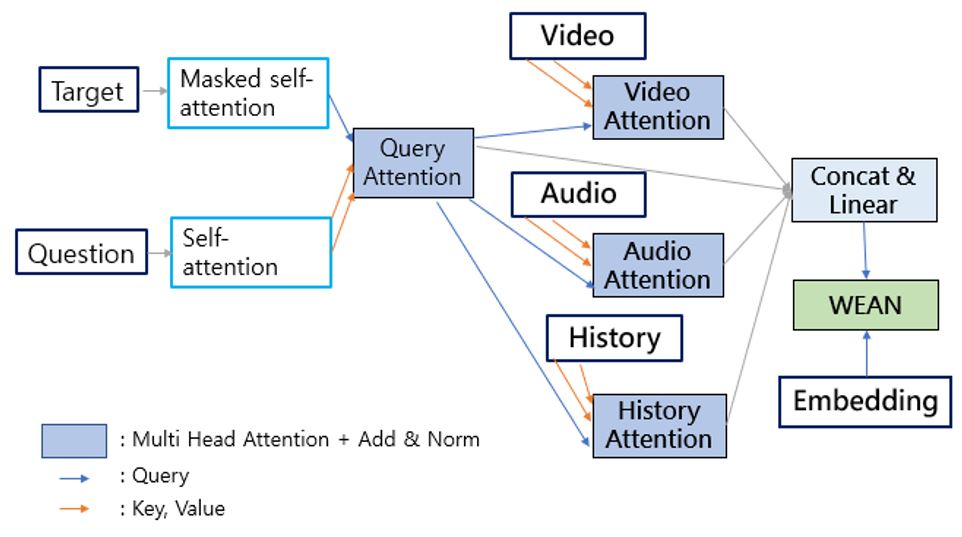}
\caption{Overall Architecture of our model}

\end{figure}

Inspired by MTN~\cite{le2019multimodal}, we propose Multimodal Semantic Transformer Network \textbf{(MSTN)} that is also based on transformer encoder-decoder framework. Our model is composed of two parts: encoder and decoder. Since we treat caption, history as different modalities, we use five modalities for this model.

\subsection{Encoder}
For the text data, we embed the source sequence to continuous representation and learn its embedding during training. We use the features extracted from i3d-flow for the video data and use the features extracted from veggies for the audio data. 
For each source sequence $X^m = (x^m_1, ...,x^m_n)$ for modality \textit{m}, we encode \(X^m\) with the encoder similar to that of transformer~\cite{vaswani2017attention} except self-attention layer which is similar to~\cite{le2019multimodal}. It is composed of positional encoding, layer normalization and feed-forward neural network. The result representation for each modality is $Z^m=(z^m_{1}, ...,z^m_n)$.

\begin{table*}[t]
\small
\centering
\begin{tabular}{L{0.25\columnwidth}C{0.1\columnwidth}C{0.1\columnwidth}C{0.1\columnwidth}C{0.1\columnwidth}C{0.1\columnwidth}C{0.1\columnwidth}C{0.1\columnwidth}C{0.1\columnwidth}C{0.1\columnwidth}C{0.1\columnwidth}C{0.1\columnwidth}C{0.1\columnwidth}}
\toprule
              & \multicolumn{4}{c}{DSTC7 (single reference)}      &  \multicolumn{4}{c}{DSTC7 (6 references)}      & \multicolumn{4}{c}{DSTC8 (6 references)}      \\
\cmidrule{2-13}
              & B-4 & M & R-L & C & B-4 & M & R-L & C & B-4 & M & R-L & C \\
\midrule
\multicolumn{13}{c}{\textbf{Text+Video}} \\
\midrule
\textbf{MTN}~\scriptsize{[1]}     & \textbf{0.135} & \textbf{0.165}  & 0.365   & \textbf{1.366} & - & -  & -   & - & - & -  & -   & - \\



\textbf{MSTN} & \textbf{0.135} & \textbf{0.165}  & \textbf{0.369}   & {1.352} & \textbf{0.377} & \textbf{0.275}  & \textbf{0.566}   & \textbf{1.115} & \textbf{0.385} & \textbf{0.270}  & \textbf{0.564}   & \textbf{1.073} \\
\textbf{MSTN w/o WEAN} & {0.131} & {0.162}  & {0.360}   & {1.298} & {0.370} & {0.271}  & {0.559}   & {1.067} & {0.376} & \textbf{0.270}  & {0.563}   & {1.040} \\

\midrule
\multicolumn{13}{c}{\textbf{Text+Video w/o caption/summary}} \\
\midrule
\textbf{S-T $\mathcal{L}_{\text{JST}}$}~\scriptsize{[2]}     & 0.115 & 0.144  & 0.335   & 1.148 & \textbf{0.382} & 0.254  & 0.537  & 1.005 & - & -  & -   & - \\
\textbf{Multi Attn}~\scriptsize{[3]}     & 0.078 & 0.113  & 0.277   & 0.727 & - & -  & -  & - & - & -  & -   & - \\
\textbf{LSTM Attn}~\scriptsize{[4]}     & 0.096 & 0.128  & 0.311   & 0.941 & - & -  & -  & - & - & -  & -   & - \\
\textbf{QG-BLSTM}~\scriptsize{[5]}     & 0.109 & 0.138  & \textbf{0.366}   & 1.132 & - & -  & -  & - & - & -  & -   & - \\
\textbf{MSTN} & \textbf{0.118} & \textbf{0.148}  & 0.348   & 
\textbf{1.211} & 0.379 & \textbf{0.261}  & \textbf{0.548}   & \textbf{1.028} & \textbf{0.375} & 0.251  & \textbf{0.544}   & \textbf{0.975} \\

\textbf{MSTN w/o WEAN} & 0.115 & 0.145  & 0.34   & 1.159 & 0.362 & 0.254  & 0.532   & 0.982 & 0.367 & \textbf{0.252}  & 0.541   & 0.967 \\

\bottomrule
\end{tabular}
\caption{
Model performance on the DSTC dataset (top scores marked in bold). 
Models [1-5] are from~\cite{le2019multimodal,hori2019joint,hori2019end,schwartz2019simple,chao2019learning}, respectively.
Note that ours* is the system that we submitted to the challenge and ours** is the system that performed additional hyperparameter tuning after the challenge.
The B-4, M, R-L, C stands for BLEU4, METEOR, ROUGE-L, and CIDEr, respectively.
}
\label{table:comparison}
\end{table*}

\subsection{Decoder}
Given the output of the encoder $Z^m = (z^m_1, ...,z^m_n)$ for each modality $m$ and target sequence $Z^y_{<t} = (z^y_1, ...,z^y_{t-1})$ the decoder generates \textit{t}-th word $y_t$. The model firstly encodes the target sequence $Z^y_{<t}$ with multi-head self-attention as in the encoder part. For video and audio data, we additionally compute question-aware attention as in \cite{le2019multimodal}, which is multi-head attention that uses a question as a query and uses other modalities as key and value. And then model computes multi-head attention with each modality $m$ as (1). Multi-Head is standard multi-head attention in transformer with 1-layer and 8-attention heads, and we use the same hyperparameter for question-aware attention. After computing attention with all of the modalities, we concatenate them and apply the linear projection. \textit{M} is the number of modalities.
\begin{equation}
\begin{aligned}
&Z^y_m = \text{MultiHead}(Z^y, Z^m, Z^m), \\
&O^y_{<t} = \text{Linear}(\text{Concat}(Z^y_\text{1}, ... , Z^y_\text{M})).
\end{aligned}
\end{equation}

\begin{figure}[t]
\centering
\includegraphics[scale=0.3]{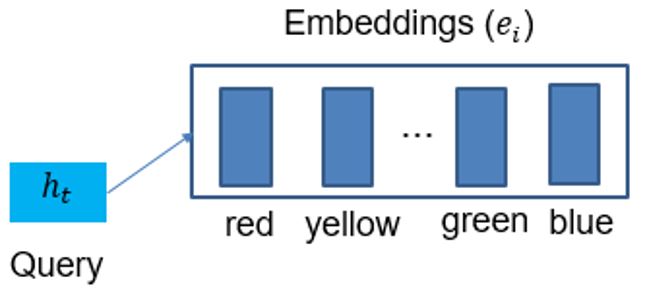}
\caption{Word Embedding Attention network}
\end{figure}

Finally, unlike the output layer of a typical decoder in sequence-to-sequence, we adopt word embedding attention network to get a score through the dot-product between $O^y_{<t}$ and embedding $e_i$ $(i = 1,2,..,n)$, where $n$ is the vocabulary size. And then take its maximum to generate the next word as follows:
\begin{equation}
\begin{aligned}
&\text{score}(o^y_\text{t-1} ,e_i) = o^y_{t-1}e^T_i, \\
&P(y_t) = \text{softmax}(\text{score}(o^y_{t-1} ,e_i)).
\end{aligned}
\end{equation}

\begin{equation}
\begin{aligned}
&\text{score}(h_\text{t} ,e_i) = h_{t}e^T_i, \\
&P(y_t) = \text{softmax}(\text{score}(h_{t} ,e_i)).
\end{aligned}
\end{equation}

\section{Experiments}

\subsection{Implementation Details}
We train the model for 20 epochs using Adam Optimizer~\cite{kingma2014adam} and use warmup steps like \cite{vaswani2017attention,le2019multimodal}. We train word embeddings with size 512 and add positional encoding to embeddings as in \cite{vaswani2017attention}. Pre-trained embedding such as GloVe\cite{pennington2014glove} does not show the difference in performance. We use only one-decoder layer and use dropout with probability 0.2 and 512 hidden units for all of the feed-forward layers. The maximum length for decoding is 30, and we use a beam search with beam size 5.

\subsection{Experimental Results}

Table~\ref{table:comparison} show the model performance on the AVSD dataset in DSTC7\cite{alamri2018audio} and DSTC8~\cite{kim2019eighth}. As the DSTC7 dataset is organized as two versions (single reference and six references), we report the model performance on both sets. Also, note that we measure the model performance with/without caption data and report the result in a separate section. To compare the model performance, we chose recently published models such as \textbf{MTN}~\cite{le2019multimodal}, \textbf{Student-Teacher $\mathcal{L}_{\text{JST}}$}~\cite{hori2019joint}, Multimodal Attention~\cite{hori2019end}, LSTM attention~\cite{schwartz2019simple}, and Question Guided BiLSTM attention~\cite{chao2019learning} that are proposed to tackle the audiovisual scene-aware task.
As shown in the Table~\ref{table:comparison}, our proposed model (\textbf{MSTN}) with retrieval style word generator outperforms the baseline model w/o it. This means that adopting a retrieval style word generator is helpful for AVSD.  Also, our model (\textbf{MSTN}) surpasses most of the other methods.

\section{Conclusion}
In this paper, we propose \textbf{MSTN}(Multimodal Semantic Transformer Networks) for audio visual scene-aware dialog. 
The proposed transformer-based encoder-decoder model employs an attention-based word embedding layer by substituting the linear projection layer in the decoder part. With this design, the model allows considering semantic information more and more when generating the answer. Experimental results show that our model with thhis design surpasses baseline models.

\section{Acknowledgments}
K. Jung is with ASRI, Seoul National University, Korea.
This work was supported by the Ministry of Trade, Industry \& Energy (MOTIE, Korea) under Industrial Technology Innovation Program (No.10073144).

\bibliographystyle{aaai}
\bibliography{paper-DSTC8.bbl}
\end{document}